\documentclass{article}
\usepackage{ijcai26}

\usepackage{times}
\usepackage{soul}
\usepackage{url}
\usepackage[hidelinks]{hyperref}
\usepackage[utf8]{inputenc}
\usepackage[small]{caption}
\usepackage{graphicx}
\usepackage{amsmath}
\usepackage{amsthm}
\usepackage{booktabs}
\usepackage{algorithm}
\usepackage{algorithmic}
\usepackage[switch]{lineno}

\usepackage{colortbl} % 색상 있는 표를 그리기 위해 필요
\usepackage{multirow} % 셀 병합을 위해 필요
\usepackage{booktabs} % 표 스타일을 세련되게 하기 위해 사용
\usepackage{array} % 가운데 정렬을 위한 패키지

\usepackage{verbatim}
\usepackage{float}
\usepackage{xspace}
\usepackage{subcaption}
\usepackage{amssymb}

\newcommand\nj[1]{\textcolor{black}{#1}}
\newcommand\ig[1]{\textcolor{black}{#1}}
\newcommand\sh[1]{\textcolor{black}{#1}}
\newcommand\igg[1]{\textcolor{black}{#1}}
\newcommand\iggg[1]{\textcolor{black}{#1}}

\newcommand\igi[1]{\textcolor{black}{#1}}
\newcommand\igii[1]{\textcolor{black}{#1}}
\newcommand\lig[1]{\textcolor{black}{#1}}
\newcommand\ligg[1]{\textcolor{black}{#1}}
\newcommand\li[1]{\textcolor{black}{#1}}
\newcommand\norm[1]{\lVert#1\rVert}
\newcommand{\eg}{\emph{e.g.}\xspace}

\title{\lig{DivCon-NeRF: Diverse and Consistent Ray Augmentation for Few-Shot NeRF}}

\author{
Ingyun Lee
\and
Jae Won Jang\and
Seunghyeon Seo\and
Nojun Kwak\thanks{The corresponding author.}\\
\affiliations
Seoul National University
\emails
\{ig.lee, pert0407, zzzlssh, nojunk\}@snu.ac.kr
}
\begin{document}

\maketitle

\begin{abstract}
Neural Radiance Field (NeRF) has shown remarkable performance in novel view synthesis but requires numerous multi-view images, limiting its practicality in few-shot scenarios. Ray augmentation has been proposed to alleviate overfitting caused by sparse training data by generating additional rays. However, existing methods, which generate augmented rays only near the original rays, exhibit pronounced floaters and appearance distortions due to limited viewpoints and inconsistent rays obstructed by nearby obstacles and complex surfaces. To address these problems, %we consider both diversity and consistency simultaneously for the first time and propose DivCon-NeRF. 
\lig{we propose DivCon-NeRF, which introduces novel sphere-based ray augmentations to significantly enhance both diversity and consistency. By employing a virtual sphere centered at the predicted surface point, our method generates \ligg{diverse} augmented rays from all 360-degree directions, facilitated by our consistency mask that effectively filters out inconsistent rays. We introduce tailored loss functions that leverage these augmentations, effectively reducing floaters and visual distortions. Consequently, our method outperforms recent few-shot NeRF approaches on the Blender, LLFF, and DTU datasets. Furthermore, DivCon-NeRF demonstrates strong generalizability by effectively integrating with both regularization- and framework-based few-shot NeRFs.} 
\end{abstract}

\section{Introduction}
\label{sec:Introduction}
\lig{Neural Radiance Field (NeRF)~\cite{mildenhall2020nerf}, a neural network-based method, has significantly advanced the performance of novel view synthesis.} However, NeRF-based models face the common deep learning challenge of performance \ig{degradation} in few-shot scenarios. A limited number of training rays with sparse viewpoints \igi{causes} overfitting. %\igg{,} \nj{and the}
\igi{Therefore,} the need for numerous images from different angles presents a substantial challenge for the practical application of NeRF-based models. To address the few-shot view synthesis problem, three main approaches are actively researched: prior-, framework-, and regularization-based methods. Each approach takes a different direction and is relatively orthogonal. Prior-based methods~\cite{yu2021pixelnerf,chen2021mvsnerf} leverage 3D knowledge through pre-training on large-scale datasets, but they incur high computational costs. Meanwhile, framework-based methods~\cite{zhu2024vanilla,zhong2024cvt} modify or extend model architectures, which can increase model complexity.

\begin{figure}[tbp]
  \centering
   \includegraphics[width=0.96\linewidth]{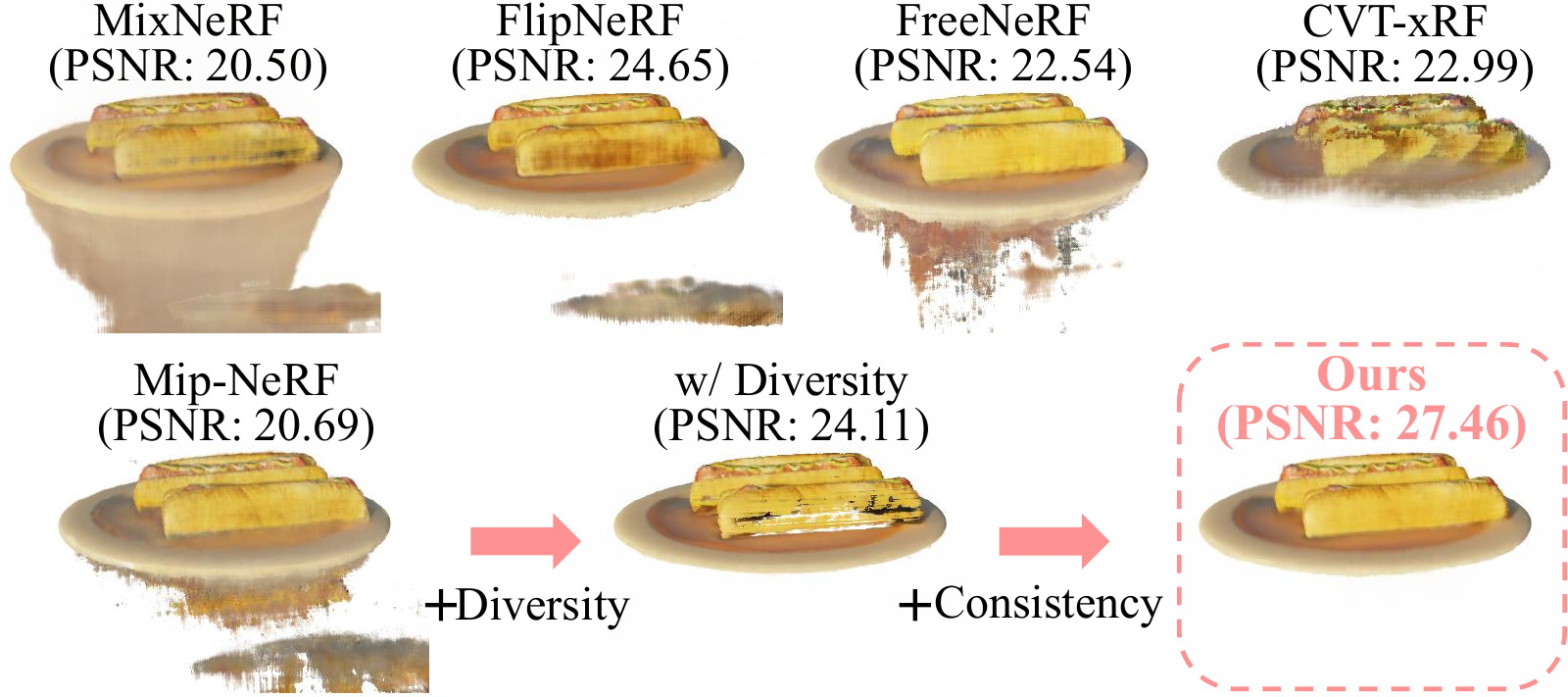}
   \caption{Rendering results of \lig{recent few-shot NeRFs} on the hotdog scene of the Blender dataset \lig{using 4 input views}. Notable floaters and appearance distortions are visible in \lig{other methods, while ours effectively addresses these artifacts by enhancing both diversity and consistency.}}
   \label{fig:start}
\end{figure}

\lig{Our proposed method is a regularization-based approach that uses explicit regularizers~\cite{xu2024few_arnerf,yang2023freenerf} or auxiliary training sources~\cite{wang2023sparsenerf}. In particular, ray augmentation generates additional rays from a limited training set, enabling the model to learn from both original and augmented rays, thereby reducing overfitting. We define two key properties for ray augmentation: ray diversity, the range of viewpoints from which augmented rays are generated, and ray consistency, the requirement that each augmented ray converges to the same surface point as its original. Existing methods~\cite{Seo_2023_CVPR,seo2023flipnerf} sample rays only near the originals to avoid occlusion. However, as shown in Fig. 1, these approaches suffer from significant floaters owing to limited viewpoint variation (low diversity) and training on inconsistent rays obstructed by nearby obstacles and complex surfaces (low consistency). Furthermore, they introduce geometric and color distortions.}

To \lig{address these limitations, }%address the limitations of existing methods, %we consider both the diversity and consistency of augmented rays simultaneously for the first time and propose DivCon-NeRF. 
\igi{we propose DivCon-NeRF, which enhances diversity while preserving consistency.} In general data augmentation, various studies~\cite{cubuk2018autoaugment,cubuk2020randaugment,xie2020unsupervised,lee2024domain,hendrycks2019augmix} \igg{have focused on} diversifying data while maintaining consistency. Based on this observation, we hypothesize that enhancing ray diversity while preserving consistency is critical for improving performance in few-shot view synthesis. \igi{To achieve this, DivCon-NeRF consists of surface-sphere and inner-sphere augmentations.} As shown in Fig.~\ref{fig:start}, our method \li{effectively} reduces floaters and appearance distortions \lig{compared to existing approaches.}
%this approach effectively reduces floaters and preserves fine object details through anti-aliasing. 

\lig{We first define a virtual 3D sphere centered at the inferred surface point, with its radius set to the distance between the camera and that point. We cast rays from both the surface and interior of the sphere toward the inferred surface point.}
 % To achieve higher-quality few-shot synthesis, we define a \igg{3D sphere} centered at the \igii{object's} inferred surface point with a radius equal to the distance between the camera and the \igii{inferred} surface point. %\nj{We then cast surface-sphere augmented rays} from the sphere's surface \nj{and} inner-sphere augmented ray\nj{s from within the sphere}, both \nj{directed} toward the inferred surface point. 
 % \igg{To be consistent with the original ray, the model should predict the same surface point for augmented rays reaching the same location.}
 %The model should predict the same surface point \nj{for augmented rays \igg{reaching the same location,} so that they are consistent with the original ray.} %as the original ray when the augmented rays are consistent. 
 In surface-sphere augmentation, a consistency mask is generated by comparing the order of high-probability surface points between the original and augmented rays, allowing the model to filter out inconsistent rays easily without relying on precise \igi{rendered} depth. \igi{We propose a ray consistency loss that aligns surface points by comparing the similarity of blending weight distributions, applying a temperature factor to assign more weight to regions with higher surface probabilities.} %We also \igi{propose} a ray consistency loss\igg{,} \nj{which} aligns surface points by comparing the similarity of the \ig{blending weight} distribution\igi{s}. \nj{A temperature factor is applied} to give more weight to regions with higher surface probabilities. 
 Additionally, \igi{we introduce a positional constraint to the bottleneck feature loss} to account for the relative positions of points. For greater diversity, inner-sphere augmentation \igii{utilizes randomized} angles and distances, providing a broader range of viewpoints. 
 
 Our \sh{DivCon}-NeRF effectively increases diversity while maintaining consistency, resulting in improved rendering quality compared to that of \lig{recent NeRF-based} methods. These results support our hypothesis that both diversity and consistency are essential for enhancing rendering performance. Our main contributions are as follows:
\begin{itemize} 
\item We experimentally demonstrate the importance of ray diversity and consistency in augmentation, analyzing their effects in both object-centric scenes and those with diverse depth ranges.
%\item We experimentally demonstrate that increasing diversity and maintaining ray consistency are crucial in ray augmentation. 
%\item \igg{We analyze how each effect varies in onbject-centric scenes and scenes with diverse depth ranges.}
\item We propose DivCon-NeRF, %a new ray augmentation method \igg{that} is the first to consider diversity and consistency simultaneously.
\igi{a novel ray augmentation method that simultaneously enhances diversity and preserves consistency.}
\item DivCon-NeRF significantly reduces floaters and appearance distortions, \lig{outperforming recent few-shot NeRFs}  \igi{on the Blender, LLFF, and DTU datasets.}
\item Our method is compatible with other regularization- and framework-based methods, demonstrating strong generalizability.
\end{itemize}

\section{Related Work}
\label{sec:Related work}
\subsection{Neural Scene Representations}
Neural scene representations have become a prominent method for encoding 3D scenes, achieving impressive results in tasks such as novel view synthesis~\cite{mildenhall2020nerf,barron2022mip360,mildenhall2022nerfdark,mueller2022instant} and 3D scene generation~\cite{jain2022zero,poole2022dreamfusion,lin2023magic3d}. These methods typically employ neural networks to model scene properties continuously, rather than relying on traditional discretized representations, such as meshes~\cite{kanazawa2018learning,liao2018deep,pan2019deep}, voxels~\cite{gadelha20173d,xie2019pix2vox,yu2021plenoxels}, or point clouds~\cite{achlioptas2018learning,thomas2019kpconv}. Neural scene representations enable highly detailed and photorealistic scene reconstructions~\cite{park2021nerfies,martin2021nerf}. Among the various approaches, NeRF~\cite{mildenhall2020nerf} has gained attention for its ability to synthesize novel views. However, NeRF-based models experience a significant \ligg{performance drop} when trained on sparse image data~\cite{zhong2024cvt,xu2024few_arnerf}, \ligg{posing a critical challenge in practical applications}. 

\subsection{Few-Shot View Synthesis}
Various studies have aimed to address few-shot view synthesis in NeRF-based models. Prior-based methods~\cite{chen2021mvsnerf,yu2021pixelnerf} pre-train models on large-scale datasets to provide prior knowledge of 3D scenes. However, these methods incur substantial costs due to \ligg{their reliance on large-scale datasets, and performance can degrade when the pre-training and target data distributions differ.}

Framework-based methods \igg{modify network structures with models such as Transformers~\cite{vaswani2017attention} and CNNs~\cite{krizhevsky2012imagenet}} and use these modified models during rendering. mi-MLP~\cite{zhu2024vanilla} incorporates input embeddings into each MLP layer for flexible learning, while CVT-xRF~\cite{zhong2024cvt} employs an in-voxel Transformer to refine radiance properties within voxels. However, these methods increase model complexity and limit integration with other framework-based methods. \igg{Additionally, appearance distortions persist due to sparse viewpoints (e.g., CVT-xRF; see Fig.~\ref{fig:start}).}

Regularization-based methods optimize each scene using regularization terms~\cite{deng2022depth} or auxiliary training sources~\cite{wang2023sparsenerf}. 
% SparseNeRF~\cite{wang2023sparsenerf} introduces a local depth ranking \igg{regularization to enhance depth consistency. }%DietNeRF~\cite{jain2021putting} employs an auxiliary semantic consistency loss using CLIP~\cite{radford2021learning} to ensure stability across views. 
\begin{comment}
DS-NeRF (참조)  utilizes depth information of sparse 3D point clouds using Structure from Motion (SfM) (참조). 
SparseNeRF (참조) introduces a robust local depth ranking from coarse depth maps.
\end{comment}
FreeNeRF~\cite{yang2023freenerf} implements \igi{a} frequency domain to mitigate artifacts. \ligg{AR-NeRF~\cite{xu2024few_arnerf} adapts rendering loss to match the frequency progression of positional encoding.} However, data scarcity remains unresolved, as these methods do not directly increase the number of training rays, leading to several floaters (e.g., FreeNeRF; see Fig.~\ref{fig:start}).
%However, data scarcity remains unresolved since these methods do not directly expand the quantity of sparse training images.

% \sh{Furthermore}, in few-shot view synthesis, 3D Gaussian Splatting (3DGS)~\cite{kerbl20233d} has received less attention than NeRF due to its reliance on initial point clouds generated from structure from motion~\cite{schoenberger2016sfm}. \igg{However, recent studies, such as DNGaussian~\cite{li2024dngaussian} and CoR-GS~\cite{zhang2024cor}, aim to address these challenges and have shown promising improvements.} Our results demonstrate that our method performs comparably with the latest few-shot 3DGS methods.
%\igi{Additionally, separate from NeRF, recent few-shot 3D Gaussian Splatting (3DGS) methods~\cite{zhang2024cor,li2024dngaussian,zhu2025fsgs} have shown considerable capabilities but face challenges in background-free scenes crucial for practical applications. We demonstrate how our method excels in these scenes.}
Additionally, recent few-shot 3D Gaussian Splatting (3DGS) methods~\cite{zhang2024cor,li2024dngaussian,zhu2025fsgs} have demonstrated \lig{promising performance in general few-shot scenarios. However, they encounter challenges in specific scenarios, such as background-free object rendering, which is crucial for practical applications like e-commerce and digital advertising.} 

%Additionally, separate from NeRF, recent few-shot 3D Gaussian Splatting (3DGS) methods have shown considerable capabilities, yet they continue to face challenges in background-free scenes. In this paper, we focus our comparisons with other few-shot NeRF methods, while the comparisons with 3DGS methods are specifically highlighted in background-free environments to illustrate their unique challenges and performance characteristics.

\begin{figure*}
  \centering
  \includegraphics[width=0.91\linewidth]{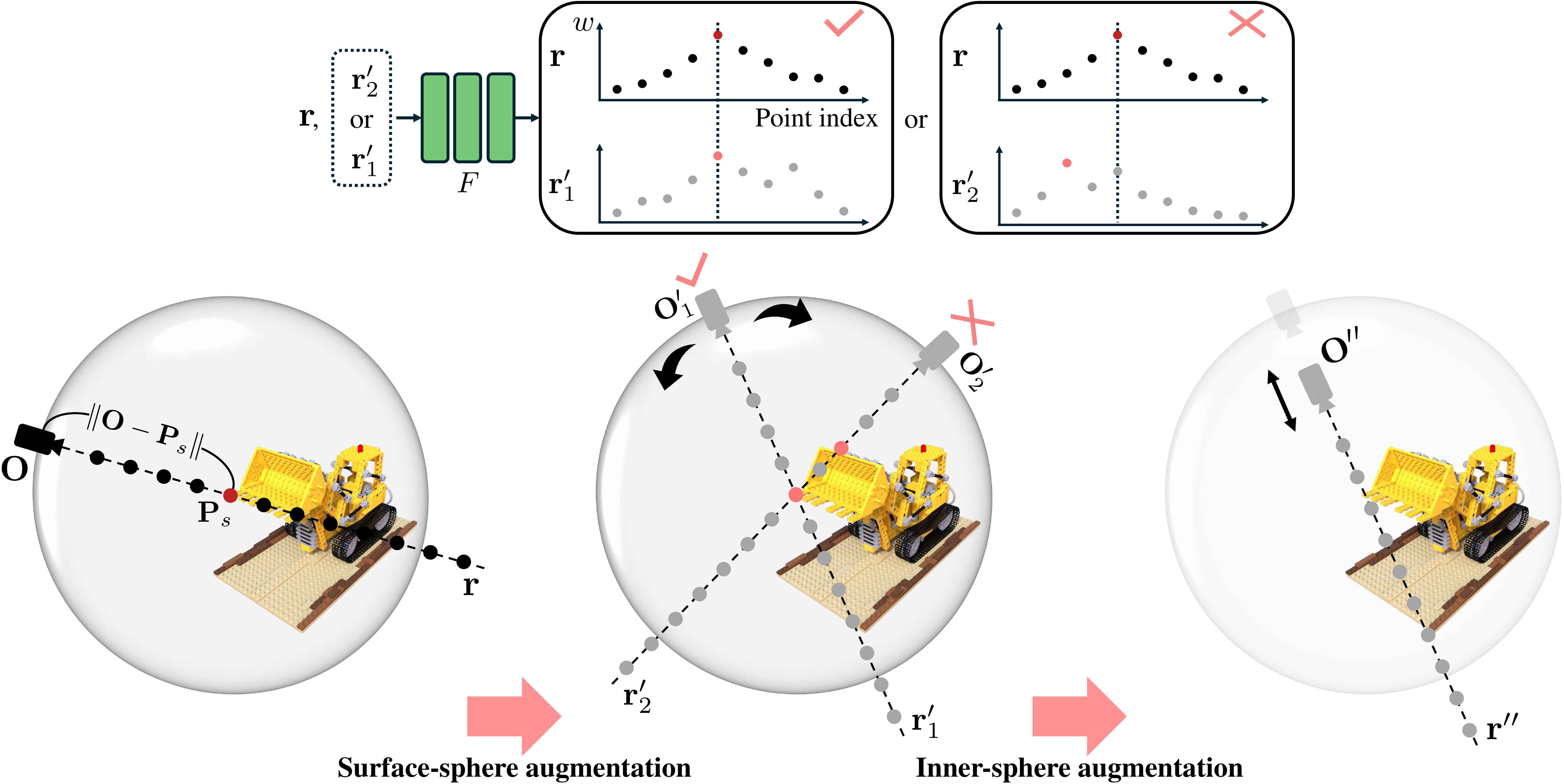}
  \caption{Overall structure of DivCon-NeRF. (Left) First, the virtual sphere is defined, centered \igg{at} the predicted surface point $\mathbf{P}_s$. (Center) A surface-sphere augmented ray $\mathbf{r}^{\prime}$ is then cast from a randomly selected position on the sphere's surface. (Right) To further enhance diversity, an inner-sphere augmented ray $\mathbf{r}^{\prime\prime}$ is generated with a randomly selected radius within the sphere and the same random angle as the surface-sphere augmented ray. (Top) To maintain consistency, the consistency mask filters out inconsistent rays (\eg, $\mathbf{r}^\prime_2$) if the sampled point with the highest blending weight does not have the same index as the original. }
  \label{fig:main}
\end{figure*}

\subsection{Ray Augmentation for Sparse Images}
Ray augmentation is a regularization-based approach that generates augmented rays from the originals. This approach~\cite{chen2022augnerf,Seo_2023_CVPR,sparf} effectively addresses the shortage of training rays by directly increasing their quantity. InfoNeRF~\cite{kim2022infonerf} reduces ray entropy to improve consistency across close views, 
% while RegNeRF~\cite{niemeyer2022regnerf} simulates unobserved viewpoints to refine the geometry and appearance.
%while Aug-NeRF~\cite{kim2022infonerf} integrates physically-grounded augmentations throughout the NeRF pipeline. 
%MixNeRF (참조) models each ray with a mixture density, enhancing depth perception. 
while FlipNeRF~\cite{seo2023flipnerf} employs flipped reflection rays to improve 3D geometry estimation. \ligg{FrugalNeRF~\cite{frugalnerf} enforces novel view consistency by jointly optimizing poses and voxel features.} \ligg{However, these methods, which augment rays only near the originals, struggle with complex surfaces and nearby obstacles, causing inconsistencies.} Furthermore, these methods generate augmented rays from limited viewpoints. \ligg{GeCoNeRF~\cite{kwak2023geconerf} applies a depth-based mask to every pixel of a warped patch rather than directly to a ray. This mask relies on image warping, which causes interpolation issues and fails to maintain equal distances from surface points to both the original and pseudo cameras for each pixel, resulting in coarse filtering. Therefore, GeCoNeRF depends on pseudo-views near the training views.} These limitations in \igi{existing methods} lead to pronounced floaters and appearance distortions due to low diversity and inadequate consistency. In contrast, our proposed method enhances both diversity and consistency, effectively addressing these limitations.

\section{Preliminaries: NeRF}
NeRF~\cite{mildenhall2020nerf} utilizes MLPs to synthesize novel views of a scene from densely sampled images. \lig{The NeRF network $F$ maps 3D coordinates $\mathbf{X}=(x,y,z)$ and viewing direction $\mathbf{d}$ to color $\mathbf{c}$ and volume density $\sigma$: }
% The input of NeRF network $F$ includes the 3D coordinates $\mathbf{X}=(x,y,z)$ of points located along a camera ray and the viewing direction $\mathbf{d}$. The output is the emitted color $\mathbf{c}$ and the volume density $\sigma$ at each point: 
\begin{equation}
    F:(\mathbf{X}, \mathbf{d}) \rightarrow(\mathbf{c}, \sigma).
    \label{eq:3.1_1}
\end{equation}
NeRF casts a single ray into 3D space for each pixel in the image. Because the ray is continuous, points along the ray are sampled. \lig{These sampled points determine the final pixel color $C$ through volume rendering.} \lig{The ray is expressed as $\mathbf{r}(t)=\mathbf{O}+t\mathbf{d}$, where $\mathbf{O}$ is the camera origin.} The predicted pixel color $C(\mathbf{r})$ is calculated as follows:   
\begin{equation}
    C(\mathbf{r})=\sum_{i=1}^N T_i\left(1-\exp \left(-\sigma_i \delta_i\right)\right) \mathbf{c}_i = \sum_{i=1}^N w_i \mathbf{c}_i,
    \label{eq:3.1_2}
\end{equation}
where $T_i=\exp \left(-\sum_{j=1}^{i-1} \sigma_j \delta_j\right)$, and \igg{$\delta_i=t_{i+1}-t_i$}. %Accumulated transmittance 
Blending weight
$w_i$ represents the contribution of the color at the $i$-th point to the pixel color $C(\mathbf{r})$. The NeRF network $F$ is optimized using a mean square error (MSE) loss to ensure that the predicted pixel color $C(\mathbf{r})$ matches the ground truth color $C_{\mathrm{gt}}(\mathbf{r})$ of the training rays:   
\begin{equation}
    \mathcal{L}_{\mathrm{MSE}}=\sum_{\mathbf{r} \in \mathcal{R}}\left\|C_{\mathrm{gt}}(\mathbf{r})-C(\mathbf{r})\right\|^2,
    \label{eq:3.1_3}
\end{equation}
\lig{where $\mathcal{R}$ represents a set of rays. However, NeRF requires dense training views and performs poorly with sparse inputs.}

\begin{comment}
\nj{논문에 지속적으로 transmittance를 이용해서 consistency mask를 만든다고 했는데 명확히는 w를 사용하기 때문에 blending 혹은 compositing weight라는 용어를 쓸 것. intro부터 고쳐야 할 것. \\
$\exp(-\sigma \delta)$: transmittance, $T:$ accumulated transmittance, $\sigma$: density (opacity), $w$: blending weight} -> 네, 알겠습니다 교수님. 다른 기호까지 전체적으로 확인하겠습니다. 
\end{comment}

\section{Method}
\label{sec:Method}

We propose DivCon-NeRF, a ray augmentation method that simultaneously enhances ray diversity and preserves ray consistency. %while preserving consistency. 
In contrast to depth-based image warping methods~\cite{kwak2023geconerf,chen2022geoaug} \lig{that require a rendering for augmentation and suffer from 
interpolation issues}, DivCon-NeRF augments rays directly in 3D space. This method %includes 
\lig{combines} surface-sphere augmentation (Sec.~\ref{subsec:surface_aug}), which generates rays on a sphere, and inner-sphere augmentation (Sec.~\ref{subsec:inner_aug}), which introduces random distance variations within the sphere, \ig{both directed toward the inferred surface point of the original ray.} 
% In surface-sphere augmentation, a progressive consistency mask filters out inconsistent rays, and two losses refine the alignment between the augmented and original rays. In inner-sphere augmentation, rays from various viewpoints are generated and utilized for robust training. 
\lig{We introduce a consistency mask and three losses tailored to these augmentations.}
The overall scheme of DivCon-NeRF is depicted in Fig.~\ref{fig:main}.

\subsection{Surface-Sphere Augmentation}
\label{subsec:surface_aug}
In contrast to the typical approach~\cite{mildenhall2020nerf}, which positions the virtual sphere at the center of the object or 3D scene to determine camera positions, we center the virtual sphere on the predicted surface point $\mathbf{P}_s$ that the original ray hits. When the original and augmented rays cast from the sphere intersect at the same surface point, the augmented ray is unobstructed. We verify the consistency of the augmented rays by comparing the surface points hit by both rays.

For surface-sphere augmentation, the sphere's center is set at $\mathbf{P}_s$, the most likely surface point
predicted by the model: $\mathbf{P}_s=\mathbf{O}+t_s  \mathbf{d}$, where $s=\operatorname{argmax}_i\left(w_i\right)$. \lig{Here, $w_i$ denotes the blending weights of the original ray $\mathbf{r}$.}
%\lig{Here, $w_i$ is derived from the original ray $\mathbf{r}$.} 
The radius is defined as $\norm{\mathbf{O}-\mathbf{P}_s}$. Consequently, the augmented ray's camera coordinates $\mathbf{O'}$ are given by:  
\begin{equation}
    \begin{gathered}
    x_{o^{\prime}}=\norm{\mathbf{O}-\mathbf{P}_s} \sin (\theta) \cos (\phi) \\
    y_{o^{\prime}}=\norm{\mathbf{O}-\mathbf{P}_s} \sin (\theta) \sin (\phi) \\
    z_{o^{\prime}}=\norm{\mathbf{O}-\mathbf{P}_s} \cos (\theta),
    \end{gathered}
    \label{eq:3.2_1}
\end{equation}
%\ig{where $(\theta,\phi)$ are spherical coordinates, identical to those used in inner-sphere augmentation, with $-\pi \leq \theta \leq \pi$ and $0 \leq \phi \leq \pi$.}
where $(\theta,\phi)$ are spherical coordinates, with $\theta \sim U[0, \pi]$ and $\phi \sim U[0, 2\pi)$. 

During training, the model progressively improves its prediction of $\mathbf{P}_s$, refining the accuracy of the virtual sphere. Notably, regardless of the point on the sphere from which an augmented ray is cast, all rays share the same distance $R$ $(= t_s  \norm{\mathbf{d}})$ and converge toward $\mathbf{P}_s$. This property of surface-sphere augmentation allows for the use of a consistency mask by estimating \lig{the index of the peak blending weight} along the ray, eliminating the need for precise depth prediction. 
\subsubsection{Consistency Mask}
To generate the consistency mask for augmented rays without constraining angle and distance, we consider all points on the virtual sphere as potential \ig{positions}
%candidates 
for $\mathbf{O'}$. In each iteration, a point on the sphere is randomly selected as $\mathbf{O'}$ for each \ig{original} ray \lig{$\mathbf{r}$}. The coordinates of the sampled points along the \ig{surface-sphere} augmented ray \lig{$\mathbf{r'}$} are computed as $\mathbf{r'} (t)=\mathbf{O'}+t\mathbf{d'}$, where $\mathbf{d'}$ %refers to
\ig{denotes the} augmented ray's camera viewing direction:    
\begin{equation}
    \mathbf{d'}=\frac{\norm{\mathbf{d}} \left(\mathbf{P}_s-\mathbf{O^{\prime}}\right)}{\norm{\mathbf{P}_s-\mathbf{O^{\prime}}}}.
    \label{eq:3.2.1_1}
\end{equation}
 The magnitude of $\mathbf{d'}$ is set equal to that of $\mathbf{d}$ to match the distribution of the original training data. The sampled points along \lig{$\mathbf{r'}$} are then fed into the model $F$. The point with the highest blending weight is identified as the most probable surface point along the augmented ray:  $\mathbf{P}_{s'}=\mathbf{O'}+t_{s'}  \mathbf{d'}$, where $s'=\operatorname{argmax}_i\left(w'_i\right)$, with the blending weight $w'_i$ derived from \lig{$\mathbf{r'}$}. \sh{Note that t}his process occurs during coarse sampling, where points are sampled at equal intervals. Because $\norm{\mathbf{P}_s-\mathbf{O}}=\norm{\mathbf{P}_s-\mathbf{O^{\prime}}}$ and $\delta=\delta'$, where $\delta$ and $\delta'$ represent the distances between adjacent samples on the original and augmented rays, respectively, $\mathbf{P}_{s'}$ of the consistent ray should be identical to $\mathbf{P}_{s}$. 
 %\nj{if it is unoccluded 교수님 이 부분은 없는게 더 낫지 않을까요?  $\mathbf{P}_{s'}$ of the augmented ray가 아니라 의도적으로 $\mathbf{P}_{s'}$ of the consistent ray라고 했습니다}. 
 To exclude occluded rays from training, we define the following consistency mask:
\igg{\begin{equation}
    M\left(\mathbf{r}, \mathbf{r}^{\prime}\right)= \begin{cases}1 & \text { if }\left|s-s^{\prime}\right|\leq\epsilon \\ 0 & \text { otherwise},\end{cases}
    \label{eq:3.2.1_2}
\end{equation}} 
\lig{where $s$ and $s'$ denote the indices of the maximum blending weights along the rays $\mathbf{r}$ and $\mathbf{r}^{\prime}$, respectively.}
% where \igg{$\mathbf{r}^{\prime}$} \lig{denotes} the \ig{surface-sphere} augmented ray %corresponding to
% \igg{derived from} the original ray $\mathbf{r}$. 
We introduce the parameter $\epsilon$ as an error tolerance, given that this process occurs during coarse sampling. \igi{If the difference between the indices exceeds $\epsilon$, the augmented ray is filtered out, indicating that the ray reaches a different surface point.}
%\ig{1. small threshold와 error tolerance 중 어떤 것이 나을까요? 말씀해주시면 앞 뒤 문장에서 동일하게 사용하겠습니다. GeConNeRF 논문 마스크에서 threshold라 언급해서 다르게 error tolerance라 언급하는 것과 + 뒤에 coarse sampling같은 얘기를 하는 것을 고려할 때 error tolerance가 더 낫지 않을까 싶습니다. 2.$s,s'$은 앞쪽에서 argmax를 통해 정의를 해주었는데 한 번 더 풀어 써주는게 더 낫다고 보시는거죠? 교수님. -> 1. 일단 통일되게 써 봐라. 2. 맞다. 다시 풀어 써 주는 게 좋을 것 같음. 아니면 argmax식에 번호를 붙인 후 식을 인용할 것.} %\nj{subscript $r$은 굳이 넣지 않아도 될 것 같은데 넣은 이유는? -> 이해 감} 
%Considering that this process occurs during coarse sampling, %and the model may not be accurate in the early iterations of training
%we introduce $\epsilon$. %\nj{시간이 지나면서 $\epsilon$을 줄이나? 워낙 작아서 계속 그대로입니다. index 차이 2이하로 설정하였습니다. and the model may not be accurate in the early iterations of training 이 부분은 혼란을 줄 수 있어 삭제했습니다.} 

 Accurately \ig{estimating} 
 \igg{$w_i$} and $w'_i$ at every point is challenging for a model during training. \igi{For example, GeCoNeRF~\cite{kwak2023geconerf}, which relies on depth for masking warped patches, employs an additional depth smoothness loss to improve rendered depth accuracy.} \li{In contrast,} determining the point with the highest blending weight via the argmax function \li{is more straightforward and effective, resulting in fewer floaters and better performance.} \ligg{This claim is empirically validated in Sec.~\ref{subsec:analysis} and the supplement.} %\igi{as demonstrated in Sec.~\ref{subsec:analysis} and the supplementary material.}    

\subsubsection{Ray Consistency Loss}
In addition to the consistency mask, we %introduce 
introduce a ray consistency loss \igi{to further align %\igg{refine} \nj{further} 
the surface points of $\mathbf{r}$ and $\mathbf{r}^{\prime}$, %the original \nj{ray} and \nj{its consistent augmented counterpart, 
enhancing overall consistency.} %\igg{We strengthen the alignment at the surface point} by \nj{increasing 
% \igi{To reinforce this alignment, we increase} the similarity between the blending weight distribution $w$ of the original ray and $w'$ of the augmented ray. Our goal is not merely to match the similarity of %the two
% \igi{these} distributions but also to ensure %the 
% \igi{their} blending weights peak at the same surface point. \igi{Augmented rays far from the original rays should not exhibit high similarity of overall distribution but should show greater similarity near the surface.} 
\lig{We apply a Kullback–Leibler (KL) divergence loss between the blending weight distributions of the original and augmented rays, using a temperature-scaled softmax.} The ray consistency loss for each ray is defined as:
\igg{\begin{equation}
    l_{\mathrm{RC}}\left(\mathbf{r}, \mathbf{r}^{\prime}\right)=K L\left(P_T \| Q_T\right), 
    \label{eq:3.2.2_1}
\end{equation}}
where
\begin{equation*}
    P_T(i)=\frac{\exp \left(w_i / T\right)}{\sum_j \exp \left(w_j / T\right)}\text { and } Q_T(i)=\frac{\exp \left(w_i^{\prime} / T\right)}{\sum_j \exp \left(w_j^{\prime} / T\right)} \text {. }
    \label{eq:3.2.2_2}
\end{equation*} 
\lig{The temperature $T$ sharpens these softmax distributions, emphasizing their peak regions. As KL divergence primarily penalizes mismatches in high-probability regions, it encourages the peak locations of these distributions to align during training. As this loss functions as a regularizer, it allows flexibility in low-probability regions. Note that augmented rays far from the originals may not exhibit strong global similarity but should exhibit consistent peak locations near the surface.} 
% The overall ray consistency loss $\mathcal{L}_{\mathrm{RC}}$ for the entire set of rays $\mathcal{R}$ is %computed
% \igi{defined} as:
% \igg{\begin{equation}
%     \mathcal{L}_{\mathrm{RC}}=\sum_{\mathbf{r} \in \mathcal{R}} M\left(\mathbf{r}, \mathbf{r}^{\prime}\right) \cdot l_{R C}\left(\mathbf{r}, \mathbf{r}^{\prime}\right).
%     \label{eq:3.2.2_3}
% \end{equation}}

% We reduce the weights of points where \igi{$w_i$} and \igi{$w_i^{\prime}$} are small\igg{er} while increasing the weights where these values are larger. To sharpen the distributions, we apply a temperature $T$. The similarity of the sharpened distributions is measured using KL divergence. %This approach results in a more precise alignment of the surface points. 

\igi{In the LLFF dataset~\cite{mildenhall2019local}, which contains many objects within a scene, augmented rays farther from the original rays tend to have more dissimilar overall distributions. \igii{For these augmented rays located beyond a certain range, we apply clipping to $w_i$ and $w_i^{\prime}$ by setting them to 0 when $i>s$, and then obtain $P_T(i)$ and $Q_T(i)$.}} 
%\igii{For these augmented rays located beyond a certain range, we apply clipping to $w_i$ and $w_i^{\prime}$ by setting them to 0 when the index $i$ exceeds $s$ and then obtain $P_T(i)$ and $Q_T(i)$.}} 
Finally, the overall ray consistency loss $\mathcal{L}_{\mathrm{RC}}$ for \lig{a} set of rays $\mathcal{R}$ is %computed
\igi{defined} as:
\igg{\begin{equation}
    \mathcal{L}_{\mathrm{RC}}=\sum_{\mathbf{r} \in \mathcal{R}} M\left(\mathbf{r}, \mathbf{r}^{\prime}\right) \cdot l_{R C}\left(\mathbf{r}, \mathbf{r}^{\prime}\right).
    \label{eq:3.2.2_3}
\end{equation}}

\igi{In contrast to InfoNeRF~\cite{kim2022infonerf}, which only applies information regularization to augmented rays with minimal viewpoint variations, our approach leverages temperature scaling and clipping to enforce a strong consistency constraint near the surface, utilizing augmented rays across all ranges for $\mathcal{L}_{\mathrm{RC}}$. Additional comparison experiments are detailed in the supplementary material. 
}

\subsubsection{Positional Bottleneck Feature Loss}
\igi{We introduce a positional constraint to the bottleneck feature loss as proposed alongside FlipNeRF~\cite{seo2023flipnerf}. In contrast to FlipNeRF, which fails to maintain a consistent positional relationship among sampled points due to the varying normal vector magnitudes, we utilize sampled points from the original and augmented rays that are equidistant from the predicted surface point. The positional bottleneck feature loss is computed using Jensen-Shannon divergence as follows:}    
\begin{equation}
    l_{\mathrm{PBF}}\left(\mathbf{r}, \mathbf{r}^{\prime}\right)=\sum_{i=1}^N \frac{1}{N} J S D\left(\psi\left(h\left(\mathbf{X}_i\right)), \psi\left(h\left(\mathbf{X}_i^{\prime}\right)\right)\right)\right.,
    \label{eq:3.2.3_1}
\end{equation}
\igi{where \igg{$\norm{\mathbf{P}_s-\mathbf{X}_i}=\norm{\mathbf{P}_s-\mathbf{X}_i^{\prime}}$}, $\psi$ denotes the softmax function, and $h(\cdot)$ maps the input to the last layer's feature space. This constraint effectively reduces geometric discrepancies between $\mathbf{X}_i$ and $\mathbf{X}_i^{\prime}$ for reliable feature matching, as demonstrated in Sec.~\ref{subsec:analysis}. Therefore, we obtain  $l_{\mathrm{PBF}}$ through the coarse sampling of surface-sphere augmentation, excluding inner-sphere augmentation. The overall positional bottleneck feature loss $\mathcal{L}_{\mathrm{PBF}}$ is defined as:}
\begin{equation} 
    \mathcal{L}_{\mathrm{PBF}}=\sum_{\mathbf{r} \in \mathcal{R}} M\left(\mathbf{r}, \mathbf{r}^{\prime}\right) \cdot l_{\mathrm{PBF}}\left(\mathbf{r}, \mathbf{r}^{\prime}\right).
    \label{eq:3.2.3_2}
\end{equation}

\subsection{Inner-Sphere Augmentation}
\label{subsec:inner_aug}
%To further enhance the diversity of augmented rays and improve the model's generalization when the camera's distance from the surface varies, both \nj{angle} and distance are randomly selected within the sphere in each iteration for each ray. 
\igi{In addition to surface-sphere augmentation, \igii{inner-sphere augmentation randomly samples the distance $R$} within the sphere in each iteration. This further enhances diversity and improves the model's robustness to variations in the camera’s distance from the surface.} %Randomness is commonly used in \nj{data augmentation to boost} diversity~\cite{cubuk2020randaugment, hendrycks2019augmix}. When %the augmented ray’s
When \igi{a pseudo} camera is positioned %\sh{locates}
% moves
farther from the surface point than the original camera, a single pixel represents a larger area, covering both the ground truth and its neighboring pixels, making it more difficult to maintain color consistency \lig{(see supplementary material for further analysis).} Therefore, %the augmented ray’s camera position is \nj{constrained} within the sphere. 
the inner-sphere augmented ray's camera coordinates $\mathbf{O''}$ are given by:  
\begin{equation}
\begin{gathered}
x_{o^{\prime\prime}}=r\norm{\mathbf{O}-\mathbf{P}_s} \sin (\theta) \cos (\phi) \\
y_{o^{\prime\prime}}=r\norm{\mathbf{O}-\mathbf{P}_s} \sin (\theta) \sin (\phi) \\
z_{o^{\prime\prime}}=r\norm{\mathbf{O}-\mathbf{P}_s} \cos (\theta),
\end{gathered}
\end{equation}
where $r \sim U(0,1]$, and $\theta$ and $\phi$ are %identical
\igi{common} \igg{random variables} %used in 
\igi{shared with surface-}sphere augmentation, %making the viewing direction the same as $\mathbf{d'}$.%can be reused for inner-sphere augmentation. 
\igi{resulting in the same viewing direction as $\mathbf{d'}$.} Therefore, we %use
apply the same consistency mask, which detects obstacles within the sphere along the line containing both augmented rays. Although this may filter out %a small number of 
\igi{a few} consistent rays, it significantly reduces the likelihood of training on inconsistent rays. 

\lig{Instead of using MSE loss,  we extend the NLL loss from MixNeRF~\cite{Seo_2023_CVPR} to our inner-sphere augmented ray $\mathbf{r}^{\prime \prime}$, enabling the model to maintain color consistency across diverse viewpoints:  }
% \lig{Instead of using MSE loss, t}he predicted color \lig{$C(\mathbf{r}^{\prime \prime})$} from \igg{the inner-sphere} augmented ray \lig{$\mathbf{r}^{\prime \prime}$} is used to maintain color consistency by employing the negative log-likelihood (NLL) loss from MixNeRF~\cite{Seo_2023_CVPR}:
\igg{
\begin{equation}
\mathcal{L}_{\mathrm{MNLL}}=-\sum_{\mathbf{r} \in \mathcal{R}} M\left(\mathbf{r}, \mathbf{r}^{\prime}\right) \cdot \log p\left(
%\mathbf{c}^{\mathrm{GT}} 
C_{\mathrm{gt}}\mid \mathbf{r}^{\prime \prime}\right),
\end{equation}}
\lig{where 
$C_{\mathrm{gt}}$ is the ground truth color of $\mathbf{r}$, and 
$p(C_{\mathrm{gt}}\mid \mathbf{r}^{\prime \prime})=\sum_{i=1}^N \pi_i \mathcal{F}(C_{\mathrm{gt}}\mid \mathbf{r}^{\prime \prime})$. Here, $\pi_i=\frac{w''_i}{\sum_{m=1}^N w''_m}$, $w''_i$ denotes the
blending weights of $\mathbf{r}^{\prime \prime}$, and $\mathcal{F}$ denotes a Laplacian distribution centered at the predicted color $C(\mathbf{r}^{\prime \prime})$.} 
%\igg{$p(C_{\mathrm{gt}}\mid\mathbf{r''})=\sum_{i=1}^N \pi''_i \mathcal{F}(C_{\mathrm{gt}} \mid \mathbf{r''})$.} \nj{Here,} $\pi''_i=\frac{w''_i}{\sum_{m=1}^N w''_m}$ with the \nj{blending weight} $w''_i$ of $\mathbf{r''}$ and $\mathcal{F}$ denotes a Laplacian distribution. } %$\mathbf{c}^{\mathrm{GT}}$ is the ground truth color of the original ray, and $\mathbf{r}_{\mathbf{r}}^{\prime \prime}$ is the inner-sphere augmented ray. 
Although the rays converge at the same surface point, the \igg{consistent} augmented ray does not necessarily have the same ground truth color as the original ray due to view-dependent effects. \lig{As the ground truth for $\mathbf{r}^{\prime \prime}$ is unavailable, we utilize the predicted color $C(\mathbf{r}^{\prime \prime})$ from coarse sampling and a probability-based NLL loss to handle view-dependent variations while maintaining color consistency. This allows the augmented ray to produce a visually consistent color that adapts naturally across views.} 

% we model the augmented ray to maintain a visually consistent color that adapts naturally across views, preserving a coherent and stable appearance.}
% %\nj{However, we model the augmented ray to maintain} consistent color, appearing stable across views to preserve a natural and coherent appearance. 
% By utilizing a probability-based NLL loss and applying this augmentation during coarse sampling, the model is encouraged to predict consistent colors with \igg{$C_{\mathrm{gt}}$} from different viewpoints, resulting in smoother renderings. 

\begin{figure}[!tbp]
    \centering
    
    % (a) 그림
    \begin{subfigure}[b]{0.45\textwidth}
        \centering
        \includegraphics[width=\textwidth]{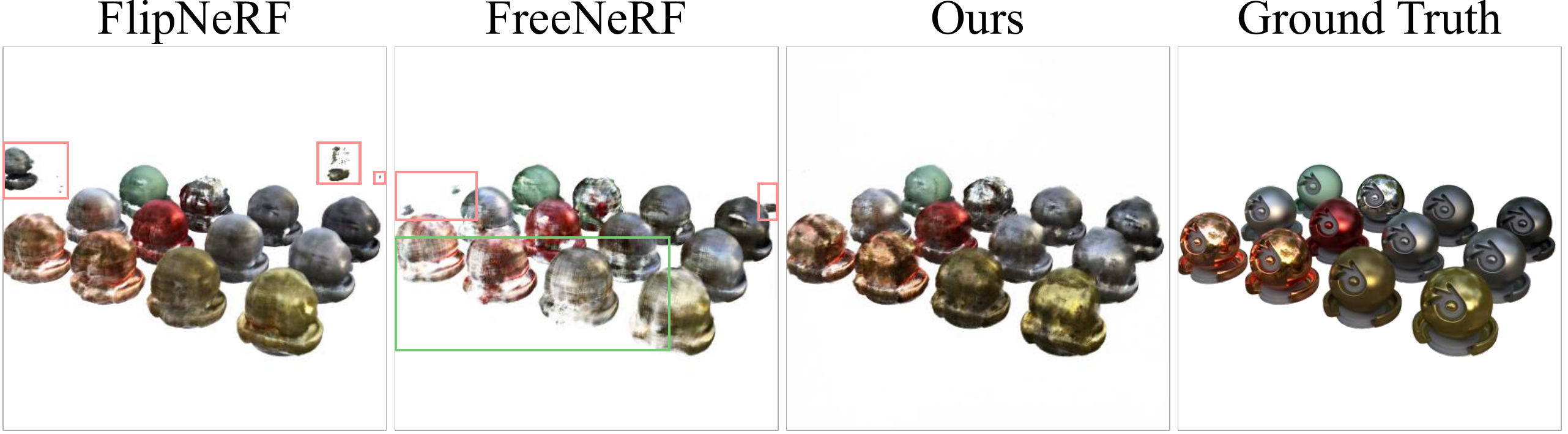}
        \caption{Blender (4-view)}
        \label{fig:qualitative_a}
    \end{subfigure}
    
    % (b) 그림
    \begin{subfigure}[b]{0.45\textwidth}
        \centering
        \includegraphics[width=\textwidth]{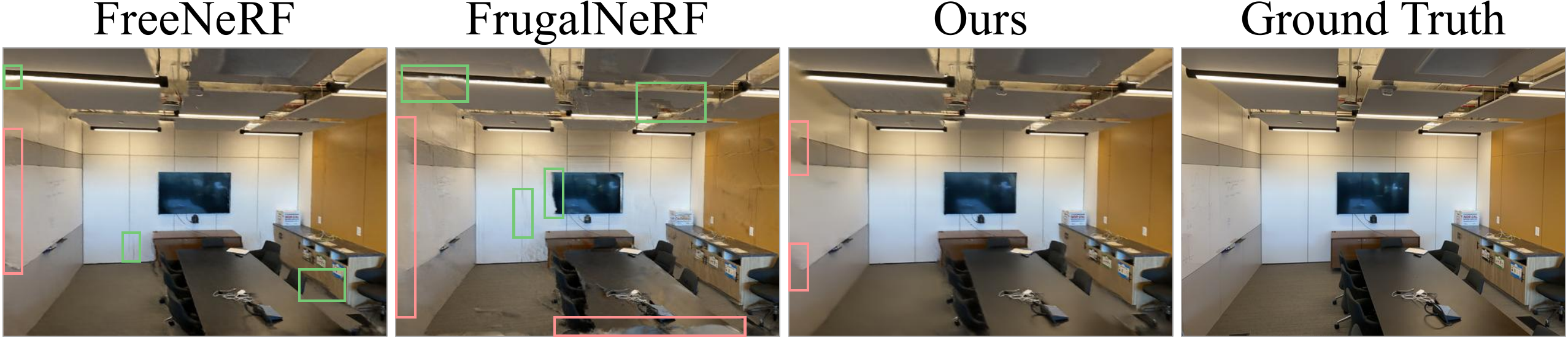}
        \caption{LLFF (3-view)}
        \label{fig:qualitative_b}
    \end{subfigure}
    
    % (c) 그림
    \begin{subfigure}[b]{0.45\textwidth}
        \centering
        \includegraphics[width=\textwidth]{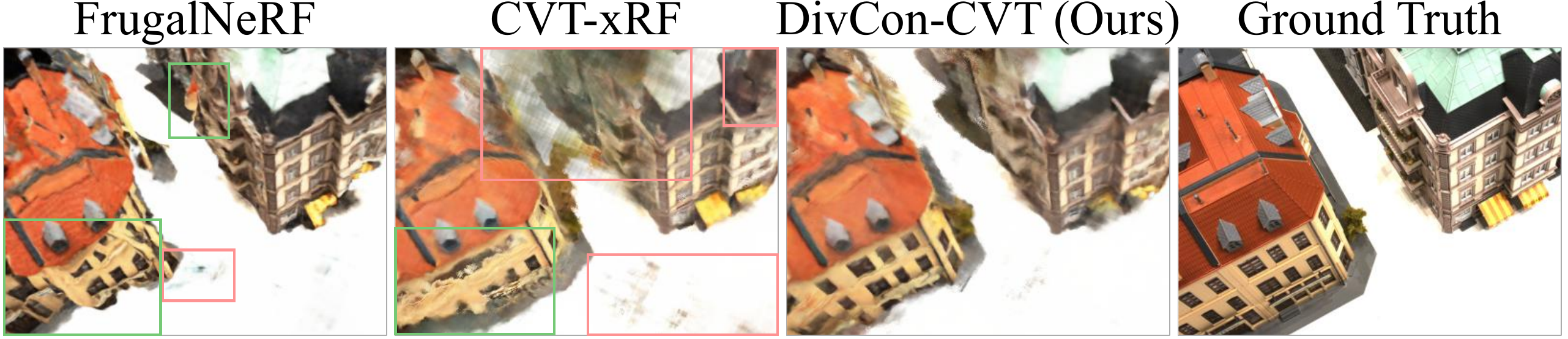}
        \caption{DTU (3-view)}
        \label{fig:qualitative_c}
    \end{subfigure}
   \caption{Qualitative comparisons on the Blender, LLFF, and DTU datasets. Red boxes indicate severe floaters, and green boxes indicate significant visual distortions. Our method exhibits fewer floaters and \igi{visual distortions than other \lig{NeRF-based} methods across all three datasets.}}
   \label{fig:qualitative}
\end{figure}

\subsection{\lig{Training Loss}}
The overall training loss is defined as follows, with $\lambda$ terms representing balancing weights for the respective losses:
\begin{equation}
\begin{aligned}
\mathcal{L} & =\mathcal{L}_{\mathrm{MSE}}+\lambda_1 \mathcal{L}_{\mathrm{RC}}+\lambda_2 \mathcal{L}_{\mathrm{PBF}} \\
& +\lambda_3 \mathcal{L}_{\mathrm{MNLL}}+\lambda_4 \mathcal{L}_{\mathrm{NLL}}+\lambda_5 \mathcal{L}_{\mathrm{UE}}.
\end{aligned}
\end{equation}
Here, \igi{$\mathcal{L}_{\mathrm{MSE}}$ is calculated using the original ray,} $\mathcal{L}_{\mathrm{NLL}}$ is the NLL loss of the original ray, introduced with MixNeRF~\cite{Seo_2023_CVPR}, and $\mathcal{L}_{\text {UE}}$ refers to the uncertainty-aware emptiness loss introduced with FlipNeRF~\cite{seo2023flipnerf}, \igi{stabilizing} the blending weights to prevent fluctuations. \igg{The supplementary material provides more details of both losses.}    
%where $\mathcal{L}_{\mathrm{NLL}}=-\sum_{\mathbf{r} \in \mathcal{R}} \log p\left(\mathbf{c}^{\mathrm{GT}} \mid \mathbf{r}\right)$, and $\mathcal{L}_{\text {UE}}$ is the uncertainty-aware emptiness (UE) loss introduced in FlipNeRF, which prevents the blending weights from fluctuating. $\mathcal{L}_{\text {UE}}$ is calculated for the original and inner-sphere augmented rays. The UE loss for a ray is calculated as: $l_{\mathrm{UE}}(\mathbf{r})=\frac{1}{N} \sum_{i=1}^N \log \left(1+\rho \cdot \eta \cdot w_i\right)$, where $\rho=\frac{1}{3} \sum_c^{\{r, g, b\}} \sum_{i=1}^N \beta_i^c$, with $\beta_i^c$ being the scale parameter of the Laplacian distribution, and $\eta$ is a hyperparameter. \ig{Experiments 전까지 내용을 전부 수정했습니다. 확인 부탁드리겠습니다.}

\section{Experiments}
\label{sec:Experiments}
\subsection{Datasets and Metrics}
 \lig{We evaluate our method against recent NeRF-based methods without external generative priors on the Blender~\cite{mildenhall2020nerf}, LLFF~\cite{mildenhall2019local}, and DTU~\cite{Jensen_2014_CVPR} datasets. For reference, we report the results of recent few-shot 3DGS methods on Blender; their performance on the LLFF and DTU datasets is reported in the supplementary material. The Blender dataset comprises 8 synthetic, background-free, object-centric scenes, while LLFF consists of 8 forward-facing real-world scenes. For the DTU dataset, we use 15 scenes following the protocol established by RegNeRF~\cite{niemeyer2022regnerf}.} %for both the DTU and LLFF datasets}. 

 % We compare our method with various SOTA approaches across the Blender~\cite{mildenhall2020nerf}, LLFF~\cite{mildenhall2019local}, and DTU~\cite{Jensen_2014_CVPR} datasets.

We report PSNR, SSIM~\cite{article_ssim}, LPIPS~\cite{zhang2018unreasonable}, and the average error (AVGE) calculated as the geometric mean of \( \text{MSE} = 10^{-\text{PSNR}/10} \), \( \sqrt{1 - \text{SSIM}} \), and LPIPS, following Niemeyer et al. (2022). %For DTU, we report the results evaluated by masked metrics, following~\cite{niemeyer2022regnerf}. 
LPIPS is evaluated using AlexNet~\cite{krizhevsky2012imagenet}, following Zhong~et~al.~\shortcite{zhong2024cvt}. Additional experimental details and extended results with more input views across Blender, LLFF, and DTU are provided in the supplement.

%Notably, previous studies did not employ standardized experimental settings for Blender dataset~\cite{mildenhall2020nerf}, which we accounted for by conducting experiments under two distinct configurations. Specifically, we sampled sparse image sets from the 100 training images in the Blender dataset using two methods. In the first method, following the approach in the MixNeRF~\cite{Seo_2023_CVPR}, we sequentially selected a subset of images to form the training dataset. In the second method, we configured the training dataset according to the sparse image selection approach described in the DietNeRF~\cite{jain2021putting}. Experiments on the LLFF~\cite{mildenhall2019local} and DTU~\cite{Jensen_2014_CVPR} datasets were conducted following the settings used in \nj{RegNeRF~\cite{niemeyer2022regnerf}}. Additionally, evaluations on the DTU dataset were conducted using metrics based on masked object images. We present quantitative evaluations of reconstruction performance by reporting PSNR, SSIM~\cite{article_ssim}, and LPIPS~\cite{zhang2018unreasonable} scores. Additionally, we compute an Average Error (AVGE)~\cite{mildenhall2020nerf}, derived from the geometric mean of \( \text{MSE} = 10^{-\text{PSNR}/10} \), \( \sqrt{1 - \text{SSIM}} \), and LPIPS. 

\renewcommand{\arraystretch}{1.0}
\begin{table}[!tbp]
\centering
\Huge
\resizebox{\columnwidth}{!}{
\begin{tabular}{l|c|c|c|c|c}
\toprule
Method & Setting & PSNR $\uparrow$ & SSIM $\uparrow$ & LPIPS $\downarrow$ & AVGE $\downarrow$ \\
\midrule
Mip-NeRF~\cite{barron2021mip} &Baseline  & 14.12 & 0.722 & 0.382 & 0.221 \\ 
\midrule

InfoNeRF~\cite{kim2022infonerf}  &\multirow{5}{*}{Regularization-based} &18.44 & 0.792 & 0.217  & 0.118 \\
RegNeRF~\cite{niemeyer2022regnerf}  & &13.71 & 0.786 & 0.339 & 0.208 \\
MixNeRF~\cite{Seo_2023_CVPR} & &18.99 & 0.807  & 0.199 & 0.113 
 \\
FlipNeRF~\cite{seo2023flipnerf} & &20.60 & 0.822  & 0.159  & 0.091  \\
FreeNeRF*~\cite{yang2023freenerf} & &20.06  & 0.815 & 0.141 & 0.092 \\
\midrule
CVT-xRF*~\cite{zhong2024cvt}& \multirow{2}{*}{Framework-based} & 19.98 & 0.819 & 0.178 & 0.096 \\
mi-MLP*~\cite{zhu2024vanilla} &   & 20.38 & 0.828 & 0.156$^{\dagger}$ & 0.084$^{\dagger}$  \\ 
\midrule
DivCon-NeRF (Ours) & \multirow{2}{*}{Regularization-based} &\underline{22.01}  &\textbf{0.843}  &\textbf{0.127}   &\textbf{0.073}  \\ 
DivCon-NeRF* (Ours) & & \textbf{22.08} & \underline{0.839}  & \underline{0.129}  & \underline{0.074}  \\ 
\midrule
\midrule
DNGaussian*~\cite{li2024dngaussian} &3DGS &19.50 & 0.808 & 0.147 & 0.093 \\
FSGS*~\cite{zhu2025fsgs} &3DGS &17.60 &0.677 &0.257 &0.140 \\ 
\bottomrule
\end{tabular}
}
\caption{Quantitative comparison on the Blender dataset \lig{using 4 views.} \lig{We report the results of NeRF-based approaches and, for reference, recent few-shot 3DGS methods (bottom).} * indicates experiments conducted under DietNeRF~\protect\cite{jain2021putting} \lig{protocols}. \lig{${\dagger}$ indicates results reported in the original paper, as the official code was not publicly available.} The best scores are in bold, and the second-best are underlined. The same notations are used in the following tables. }
\label{tab:blender}
\end{table}

\renewcommand{\arraystretch}{1.0}
\begin{table}
\centering
\Huge
\resizebox{\columnwidth}{!}{
\begin{tabular}{l|c|cccc}
\toprule
\multirow{1}{*}{Method} & \multirow{1}{*}{\centering Setting} &  PSNR $\uparrow$ & SSIM $\uparrow$ & LPIPS $\downarrow$ & AVGE $\downarrow$ \\ 
\midrule
Mip-NeRF~\cite{barron2021mip} & {Baseline}  & 15.53 & 0.363 & 0.445 & 0.223 \\ 
\midrule
SRF-ft~\cite{srf} &\multirow{3}{*}{Prior-based} &17.07 &0.436 &0.496 &0.198 \\ 
PixelNeRF-ft~\cite{yu2021pixelnerf} &  &  16.17 & 0.438 & 0.473 & 0.210 \\
MVSNeRF-ft~\cite{chen2021mvsnerf}  & &  17.88 & 0.584 & 0.260 & 0.143 \\
\midrule
% DietNeRF~\cite{jain2021putting} &\multirow{7}{*}{Regularization-based} &14.94 & 0.370 & 0.423 & 0.221  \\
RegNeRF~\cite{niemeyer2022regnerf} &\multirow{5}{*}{Regularization-based} &19.08 & 0.587 & 0.263 & 0.132 \\
% DS-NeRF~\cite{deng2022depth} &\multirow{5}{*} &18.12 &0.563 &0.269 &0.145 \\
FlipNeRF~\cite{seo2023flipnerf} & & 19.34 & 0.631 & 0.235 & 0.123 \\
FreeNeRF~\cite{yang2023freenerf} & &19.63 & 0.612 & 0.240 & 0.122 \\
AR-NeRF~\cite{xu2024few_arnerf} && \underline{19.90} & \underline{0.635} & 0.283$^{\dagger}$ & 0.126$^{\dagger}$ \\
FrugalNeRF~\cite{frugalnerf} & &19.49 & 0.621 & \underline{0.229} & \underline{0.121}  \\
\midrule
mi-MLP~\cite{zhu2024vanilla}& \multirow{1}{*}{Framework-based} & 19.75 & 0.614 & 0.300$^{\dagger}$ & 0.125$^{\dagger}$ \\ 
\midrule
DivCon-NeRF (Ours) & Regularization-based & \textbf{20.46} &\textbf{0.662} &\textbf{0.221} & \textbf{0.109} \\ 
\bottomrule
\end{tabular}
}
\caption{Quantitative comparison on the LLFF dataset under the 3-view setting. Our method achieves \igi{the best performance among \ligg{all NeRF-based} methods.} We report the fine-tuning (ft) performance for prior-based methods.}
\label{tab:llff}
\end{table}

\subsection{\lig{Comparisons}}
%We compare our method with various SOTA approaches across the Blender, LLFF, and DTU datasets. 
%\nj{Table \ref{tab:blender},  \ref{tab:llff}, \ref{tab:dtu} compare our DivCon-NeRF with} various \nj{SOTA approaches of few-shot rendering,} not limited to ray augmentation methods. %\nj{They show the superiority of our method.} 

\subsubsection{Blender}
Our method significantly outperforms all compared \lig{NeRF-based} approaches on the Blender dataset, as shown in Tab.~\ref{tab:blender}. For a fair comparison, we follow the experimental protocols of MixNeRF~\cite{Seo_2023_CVPR} (no mark) and DietNeRF~\cite{jain2021putting} (marked by $^*$). In both cases, our method consistently achieves the highest performance \lig{by a substantial margin.} As illustrated in Fig.~\ref{fig:qualitative_a}, it effectively reduces floaters and appearance distortions compared to \lig{existing methods}, thereby reinforcing the importance of consistency and diversity, as hypothesized. For completeness, we include the results of recent few-shot 3DGS methods~\cite{zhu2025fsgs,li2024dngaussian} at the bottom of Tab.~\ref{tab:blender}; these perform notably worse on background-free, object-centric scenes (see Sec.~\ref{Effectiveness in Background-Free Scenes} and the supplementary material).

\subsubsection{LLFF}
 As shown in Tab.~\ref{tab:llff}, our method achieves a substantial performance improvement over other approaches, \lig{demonstrating its effectiveness across both synthetic and real-world datasets.} %Considering diversity and consistency, our model performs well even in scenes with varying depth. 
 As shown in Fig.~\ref{fig:qualitative_b}, our \igi{method} more effectively prevents geometric distortions compared with other \igi{methods} owing to enhanced consistency, better preserving the continuity of straight lines on wall surfaces.

\renewcommand{\arraystretch}{1.0}
\begin{table}
\centering
\Huge
\resizebox{\columnwidth}{!}{
\begin{tabular}{l|c|cccc}
\toprule
\multirow{1}{*}{Method} & \multirow{1}{*}{\centering Setting} &  PSNR $\uparrow$ & SSIM $\uparrow$ & LPIPS $\downarrow$ & AVGE $\downarrow$ \\ 
\midrule
Mip-NeRF~\cite{barron2021mip} & Baseline  & 9.02 & 0.570 & 0.339 & 0.309 \\ 
\midrule
SRF-ft~\cite{srf} &\multirow{3}{*}{Prior-based} &15.68 &0.698 &0.260 &0.160 \\ 
PixelNeRF-ft~\cite{yu2021pixelnerf} &   &  18.95 & 0.710 & 0.242 & 0.121 \\
MVSNeRF-ft~\cite{chen2021mvsnerf}  & &  18.54 & 0.769 & 0.168 & 0.107 \\
\midrule
% DS-NeRF~\cite{deng2022depth} &\multirow{6}{*}{Regularization-based} &19.48 &0.695 &0.269 &0.121 \\
MixNeRF~\cite{Seo_2023_CVPR}  &\multirow{5}{*}{Regularization-based} &18.95 & 0.744 & 0.203 & 0.113 \\
FlipNeRF~\cite{seo2023flipnerf} & &19.55 & 0.767 & 0.180 & 0.101 \\
FreeNeRF~\cite{yang2023freenerf} & &19.92 & 0.787 & 0.147 & 0.092\ \\
AR-NeRF~\cite{xu2024few_arnerf} && 20.36 & 0.788 & 0.187$^{\dagger}$ & 0.095$^{\dagger}$ \\
FrugalNeRF~\cite{frugalnerf} & &\underline{21.06} & 0.792 & 0.160 & 0.087  \\
\midrule
CVT-xRF~\cite{zhong2024cvt} & \multirow{1}{*}{Framework-based} & 21.00 & \underline{0.837} & \underline{0.132} & \underline{0.077} \\
\midrule
DivCon-Flip (Ours) & (Reg.+Reg.)-based & 19.95 & 0.769 & 0.163 & 0.096 \\ 
DivCon-CVT (Ours) & (Reg.+Frame.)-based &\textbf{22.01} & \textbf{\iggg{0.858}} &\textbf{\iggg{0.130}} & \textbf{ \iggg{0.069}} \\ 
\bottomrule
\end{tabular}
}
\caption{Quantitative comparison on the DTU under the 3-view setting. \lig{We evaluate our method's integration with existing approaches.} Results are evaluated by masked metrics following \ligg{Niemeyer~et~al.~\protect\shortcite{niemeyer2022regnerf}.}}
\label{tab:dtu}
\end{table}

\subsubsection{DTU}
To demonstrate the generalizability of our method, we integrate it with the regularization-based FlipNeRF~\cite{seo2023flipnerf} and the framework-based CVT-xRF~\cite{zhong2024cvt}. As shown in Tab.~\ref{tab:dtu}, our DivCon-Flip outperforms FlipNeRF \igg{in all metrics}, \igi{and} our DivCon-CVT achieves \lig{the best} performance. %\igi{We observe improved LPIPS performance in DivCon-Flip while similar LPIPS performance in DivCon-CVT. %because CVT-xRF~\cite{zhong2024cvt}, unlike FlipNeRF~\cite{seo2023flipnerf}, already utilizes voxel-based consistency. 
Integrating our method with CVT-xRF substantially improves PSNR, SSIM, and AVGE while closely maintaining LPIPS. \lig{This suggests that our method is architecture-agnostic and effectively complements framework-based methods.} As illustrated in Fig.~\ref{fig:qualitative_c}, our method yields cleaner renderings than other methods.

\subsection{Analysis}
\label{subsec:analysis}

\subsubsection{Ray Diversity} 
\begin{figure}[!tbp]
  \centering
   \includegraphics[width=0.95\linewidth]{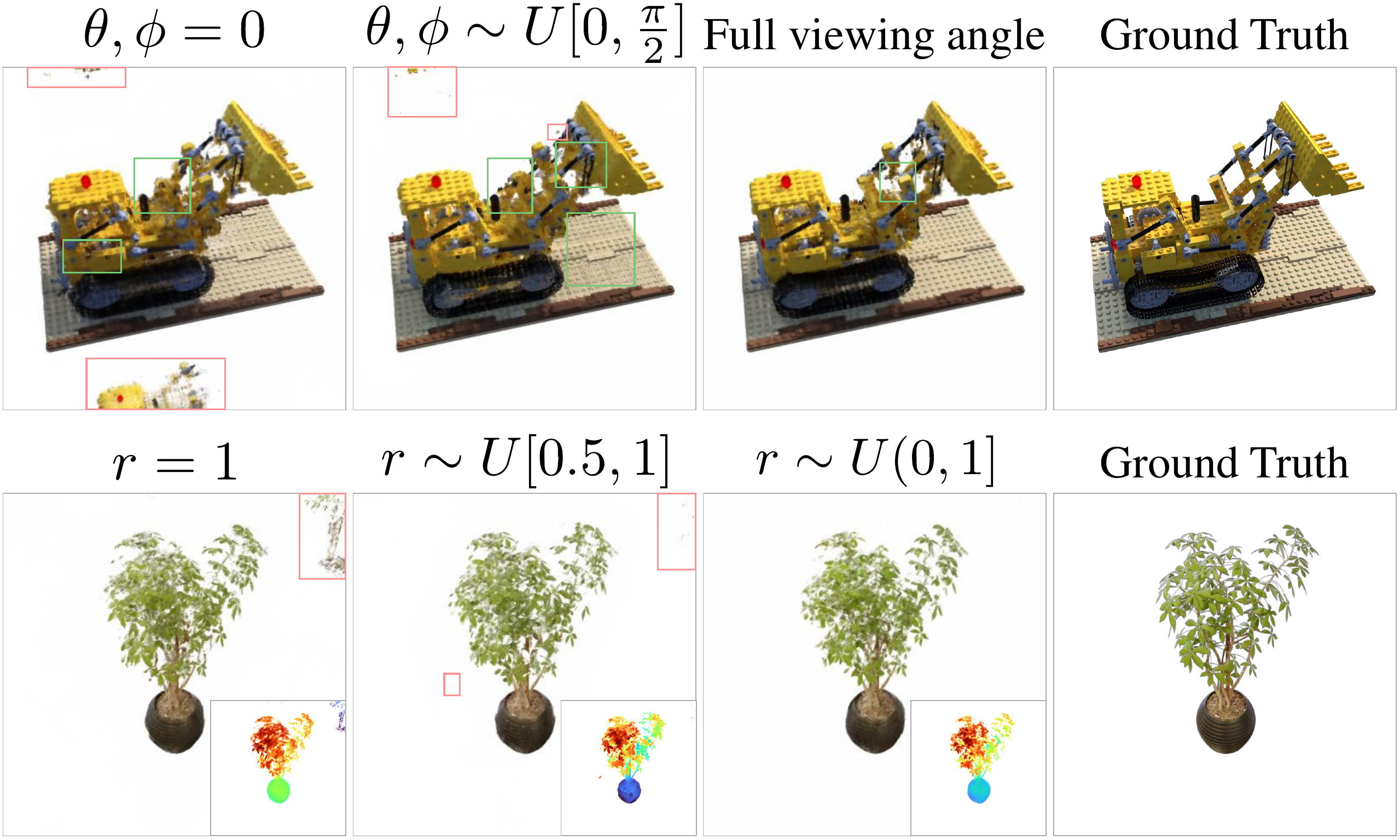}
   \caption{Analysis of diversity for angle and radius on Blender. \igi{The} insets in the second row show depth maps. Best viewed \igg{when} enlarged.}
   \label{fig:diversity_in_angle}
\end{figure}

We evaluate the impact of ray diversity on rendering quality by progressively expanding the ranges of $\theta$, $\phi$, and $r$. As shown in Fig.~\ref{fig:diversity_in_angle}, low diversity results in severe floaters (red boxes), while increased diversity notably reduces these artifacts, indicating the importance of augmented rays from diverse viewpoints. In particular, when the angular range is narrow, more floaters and pronounced visual distortions are observed (green boxes). However, employing a full viewing angle without the consistency mask can lead to %\igg{losing} finer details due to excessive sampling and
challenges in maintaining ray consistency.

%\subsection{Analysis of consistency}
%\label{subsec:analysis of consistency}

\subsubsection{Absence of Consistency Mask}
\begin{figure}[!tbp]
    \centering
    
    % (a) 그림
    \begin{subfigure}[b]{0.45\textwidth}
        \centering
        \includegraphics[width=\textwidth]{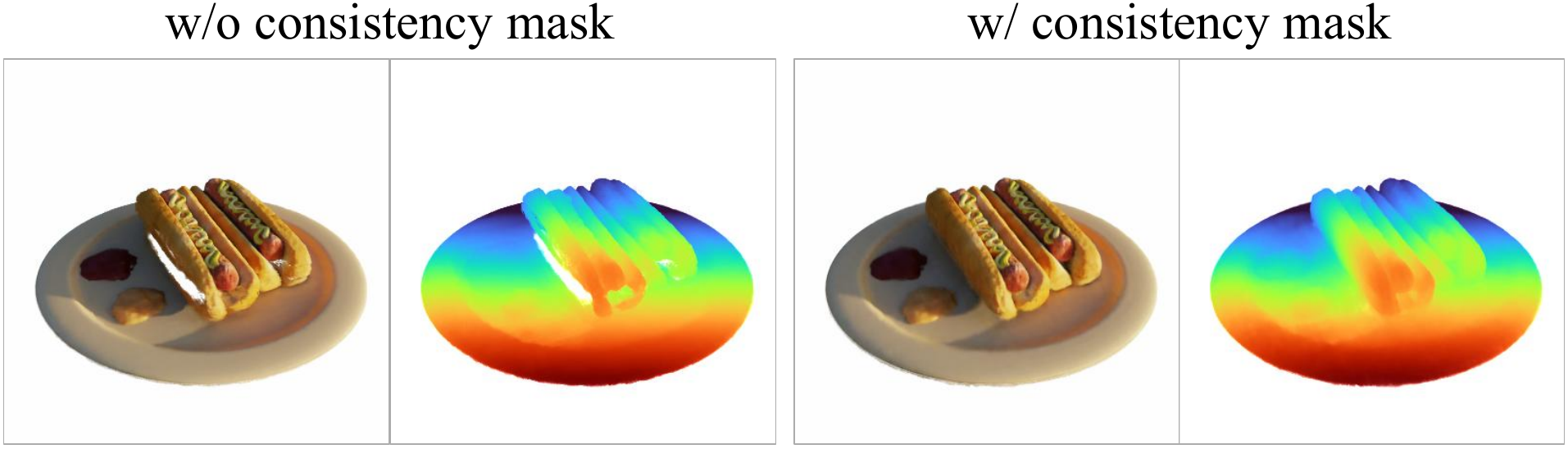}
        \caption{Blender (4-view)}
        \label{fig:absence_of_sphere_augmentation_a}
    \end{subfigure}
    
    % (b) 그림
    \begin{subfigure}[b]{0.45\textwidth}
        \centering
        \includegraphics[width=\textwidth]{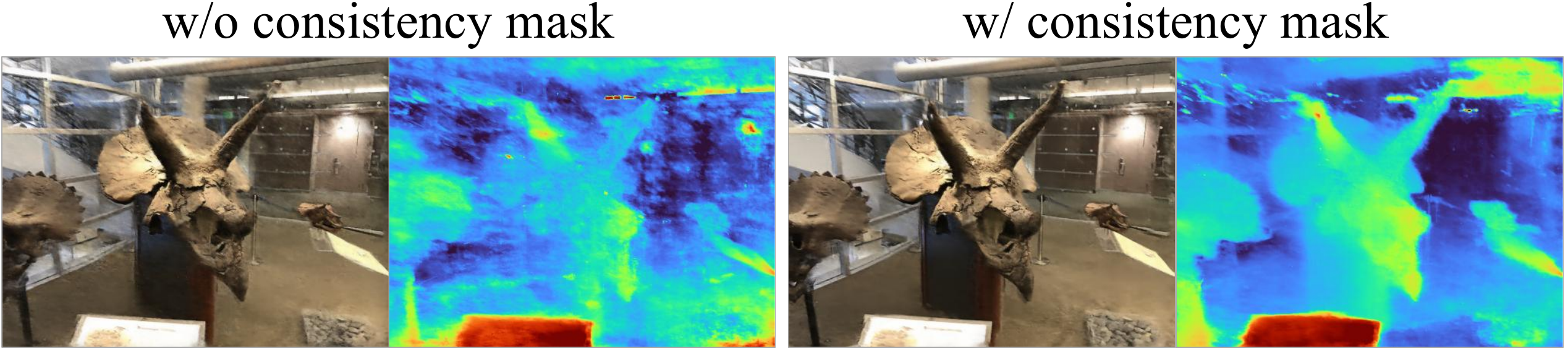}
        \caption{LLFF (3-view)}
        \label{fig:absence_of_sphere_augmentation_b}

    \end{subfigure}
   \caption{Comparison of \lig{rendered RGB and depth images} with and without the consistency mask.}
\end{figure}

To analyze the importance of consistency, we evaluate our method without the consistency mask. In object-centric scenes, as shown in Fig.~\ref{fig:absence_of_sphere_augmentation_a}, reduced consistency leads to significant visual distortions. The corresponding depth map reveals empty regions on the object surfaces, indicating that the rays fail to locate the surface. This suggests that inconsistent rays are used during training, increasing uncertainty in depth estimation and resulting in unreliable depth predictions. This highlights the crucial role of consistency in ray augmentation for preventing visual distortions. As shown in Fig.~\ref{fig:absence_of_sphere_augmentation_b}, diversity alone \igg{cannot} eliminate many floaters in the LLFF scene with varying depth\igg{s}. In this case, consistency masks effectively reduce floaters and appearance distortions. Therefore, \igi{our model demonstrates robust performance across both synthetic and real-world scenes by enhancing diversity and consistency.}

\begin{comment}
\begin{table}[t]
\centering
\renewcommand{\tabcolsep}{0.8mm}
\begin{tabular}{c|cccc|ccc}
\toprule
& Base & $\mathcal{L}_{\mathrm{RC}}$ & $\mathcal{L}_\text{PBF}$ & $\mathcal{L}_\text{MNLL}$ & {PSNR $\uparrow$} & {SSIM $\uparrow$} & {LPIPS $\downarrow$} \\
\midrule
(1) & \checkmark & & & &  18.86 & 0.803 & 0.212 \\
\midrule
(2) & \checkmark & \checkmark & & & 19.60 & 0.802 & 0.182  \\
(3) & \checkmark & \checkmark & \checkmark & &  20.44 & 0.814 & 0.097  \\
(4) & \checkmark & \checkmark & \checkmark & \checkmark &  22.08 & 0.839 &0.129\\
\bottomrule
\end{tabular}
\vspace{-2mm}
\caption{
\textbf{Ablation study.}
}
\label{tab:loss_components}
\end{table}
\end{comment}

\renewcommand{\arraystretch}{0.8}
\begin{table}[!tbp]
\renewcommand{\tabcolsep}{1.1mm}
\begin{tabular}{l|ccc}
\toprule
Methods & PSNR $\uparrow$ & SSIM $\uparrow$ & LPIPS $\downarrow$ \\
\midrule
DivCon-NeRF (w/ argmax) & \textbf{20.46} & \textbf{0.662} & \textbf{0.221}  \\ 
DivCon-NeRF (w/ depth) & 19.25  & 0.607 & 0.289 \\ 
\bottomrule
\end{tabular}
\caption{Comparison of using the argmax function and \igi{rendered} depth on LLFF.}
\label{tab:direct_use_of_depth}
\end{table}

\subsubsection{\igi{Limitations of Using Rendered Depth}}
%\paragraph{\igi{Comparison of using rendered depth.}}

\lig{We compare the performance of our consistency mask when using the argmax function versus rendered depth.} Using \igi{rendered} depth results in lower performance, as shown in Tab.~\ref{tab:direct_use_of_depth}. This suggests, that because the depth calculation depends on accurate blending weights for all points, the argmax-based masking is more effective for maintaining consistency.

%\subsection{Ablation study}
%\label{subsec:Ablation study}
\subsubsection{Ablation Study}
Tab.~\ref{tab:loss_components} \lig{shows} the impact of each component in our method. \igii{In (2), adding only $\mathcal{L}_{\mathrm{RC}}$ to align surface points shows a notable performance improvement, except for SSIM. As shown in (3), incorporating $\mathcal{L}_\text{PBF}$ significantly surpasses the baseline across all metrics. Adding $\mathcal{L}_\text{MNLL}$ of inner-sphere augmented rays achieves the ideal performance in (4), underscoring the importance of enhancing diversity while preserving consistency.} %\igg{The decrease in LPIPS performance can be easily addressed by additionally introducing a bottleneck feature loss for the inner-sphere augmented rays.}

\igii{In (5), $\mathcal{L}_\text{BF}$ represents the bottleneck feature loss %, which does not consider the relative positions of points. 
\igi{proposed with FlipNeRF~\cite{seo2023flipnerf}.} Substituting $\mathcal{L}_\text{PBF}$ with $\mathcal{L}_\text{BF}$ results in substantial overall performance degradation, indicating the importance of incorporating relative positional information for reliable feature matching.}

% \paragraph{Effectiveness in background-free scenes}
\subsection{\igi{Effectiveness in Background-Free Scenes}}
\label{Effectiveness in Background-Free Scenes}

\begin{figure}[!tbp]
    \centering
    \includegraphics[width=0.98\linewidth]{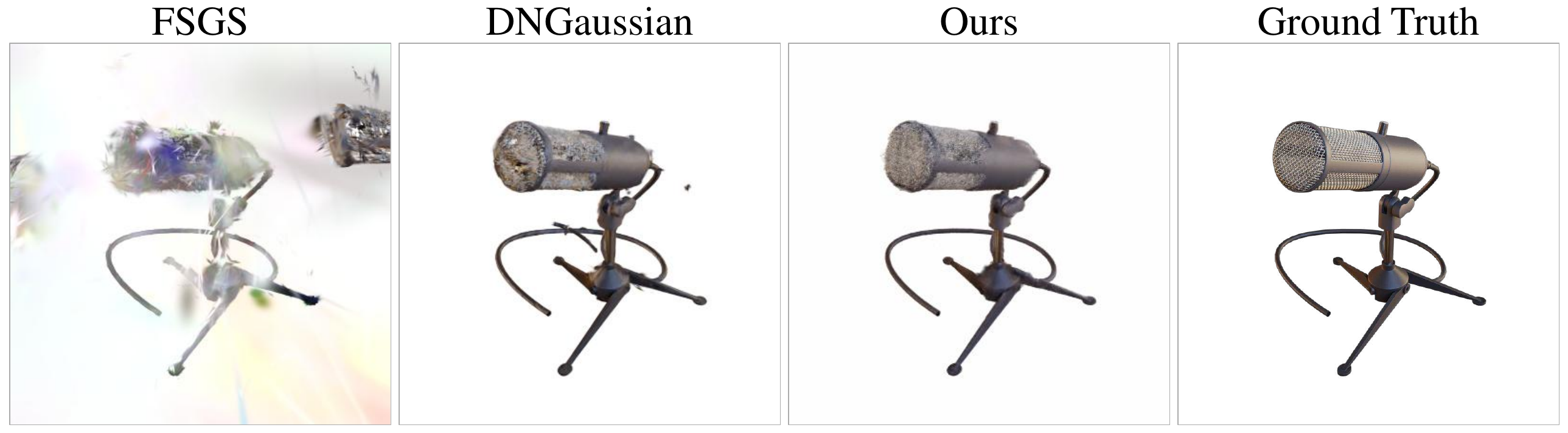}
  \caption{Qualitative comparison of our method with recent few-shot 3DGS methods on Blender in the 4-view setting.}
   \label{fig:effectiveness in background-free scenes}
   
\end{figure}

Our method is highly effective in background-free, object-centric scenes. In contrast to recent few-shot 3DGS methods~\cite{zhu2025fsgs,li2024dngaussian} which suffer from significant distortions and floaters (Fig.~\ref{fig:effectiveness in background-free scenes}, Tab.~\ref{tab:blender}), our approach produces cleaner renderings with no floaters. This makes it especially suitable for applications, such as e-commerce and digital advertising, where accurate multi-view rendering of isolated products is essential for flexible scene composition. \lig{Additional analysis on Blender and other scene types is provided in the supplementary material.}

% Fig.~\ref{fig:effectiveness in background-free scenes} \lig{and Tab.~\ref{tab:blender}} show that even recent few-shot 3DGS methods~\cite{zhu2025fsgs, li2024dngaussian} still suffer from severe distortions and floaters. 

\renewcommand{\arraystretch}{0.1}
\begin{table}[t]
\centering
\renewcommand{\tabcolsep}{0.4mm}
\begin{tabular}{c|cccc|c|ccc}
\toprule
& Base & $\mathcal{L}_{\mathrm{RC}}$ & $\mathcal{L}_\text{PBF}$ & $\mathcal{L}_\text{MNLL}$ & $\mathcal{L}_\text{BF}$ & {PSNR $\uparrow$} & {SSIM $\uparrow$} & {LPIPS $\downarrow$} \\
\midrule
(1) & \checkmark & & & & & 18.86 & 0.803 & 0.212 \\
\midrule
(2) & \checkmark & \checkmark & & & & 19.60 & 0.802 & 0.182  \\
(3) & \checkmark & \checkmark & \checkmark & & &  20.69 & 0.820 & 0.168  \\
(4) & \checkmark & \checkmark & \checkmark & \checkmark & & \textbf{22.08} & \textbf{0.839} & \textbf{0.129}\\
\midrule
\midrule
(5) & \checkmark & \checkmark & & \checkmark & \checkmark & 20.73 & 0.817 &0.149 \\
\bottomrule
\end{tabular}
\caption{
\lig{Ablation study. Base includes $\mathcal{L}_{\mathrm{NLL}}$ and $\mathcal{L}_{\mathrm{UE}}$. We highlight the effect of the newly introduced loss terms.}
}
\label{tab:loss_components}
\end{table}

\section{Conclusion}
\label{sec:Conclusion}
In this paper, we present DivCon-NeRF, which effectively reduces floaters and visual distortions in \lig{NeRF-based} few-shot view synthesis. %We are the first to enhance diversity and preserve consistency simultaneously
% \igi{Due to the fine filtering of inconsistent augmented rays by our consistency mask, we are the first to utilize pseudo views from all 360-degree directions based on a virtual sphere. 
\lig{Our consistency mask effectively filters out inconsistent augmented rays, making us the first to utilize pseudo views from all 360-degree directions based on a virtual sphere.} Our method simultaneously enhances diversity and preserves consistency, with experimental results confirming its effectiveness in both \lig{object-centric scenes and real-world environments.} \lig{Our results demonstrate that ray diversity and consistency are crucial for few-shot NeRF.}

\section*{Acknowledgments}
This work was supported by the Korean Government through the grants from IITP (RS-2021-II211343 and RS-2025-25442338) and KOCCA (RS-2024-00398320).

\bibliographystyle{named}
\bibliography{ijcai26}

\end{document}